\documentclass[letterpaper, 10 pt, conference]{ieeeconf}  

\IEEEoverridecommandlockouts                              
\usepackage[l2tabu,orthodox]{nag}

\usepackage[
    backend=bibtex8,
    style=ieee,
    sorting=none,
    natbib=true,
    doi=false,
    isbn=false,
    url=false,
    eprint=false,
    maxcitenames=1,
    mincitenames=1
]{biblatex}

\usepackage[pdftex,colorlinks]{hyperref}

\usepackage[printonlyused]{acronym}

\usepackage{siunitx}
\usepackage[all]{nowidow}

\usepackage[dvipsnames]{xcolor}

\usepackage{lipsum}

\usepackage{xspace} 
\newcommand{\ie}{i.e.,\xspace{}}

\usepackage[pdftex]{graphicx}

\usepackage{epstopdf}

\usepackage{import}

\graphicspath{{./latexGoodPractices/}}

\usepackage{booktabs}

\usepackage{tabularx}
\usepackage{multirow, multicol}

\usepackage{amssymb,amsfonts,amsmath,amscd}

\usepackage{bm}

\newcommand{\bbm}{\begin{bmatrix}}
\newcommand{\ebm}{\end{bmatrix}}

\usepackage[normalem]{ulem}

\title{\LARGE \bf 
    Field Report on a Wearable and Versatile Solution for Field Acquisition and Exploration
}

\author{Olivier Gamache$^{1}$, Jean-Michel Fortin, Mat\v ej Boxan, François Pomerleau, Philippe Giguère$^{1}$
\thanks{*This research was supported by Fonds de Recherche du Québec Nature et technologies (FRQNT) Team grant 254912 and Natural Sciences and Engineering Research Council of Canada (NSERC) DRC Grant through the grant CRD 538321-18, in collaboration with FP Innovations and Resolute Forest Products.}
	\thanks{$^{1}$Northern Robotics Laboratory, Université Laval, Québec City, Québec, Canada
		{\texttt{\small olivier.gamache@norlab.ulaval.ca}} and \texttt{\small {philippe.giguere@ift.ulaval.ca}}}%
}

\usepackage[switch]{lineno}
\usepackage[font=small]{caption}
\usepackage{subcaption}

\acrodef{SLAM}{Simultaneous Localization And Mapping}
\acrodef{VSLAM}{Visual Simultaneous Localization And Mapping}
\acrodef{VO}{Visual Odometry}
\acrodef{AE}{Automatic-Exposure}
\acrodef{CRF}{Camera Response Function}
\acrodef{DN}{Digital Number}
\acrodef{SNR}{Signal-to-Noise Ratio}
\acrodef{HDR}{High Dynamic Range}
\acrodef{IMU}{Inertial Measurement Unit}
\acrodef{FPS}{frames per second}
\acrodef{GNSS}{Global Navigation Satellite System }
\acrodef{API}{Application Programming Interface}
\acrodef{GP}{Gaussian Process}
\acrodef{RMSE}{Root-Mean-Square Error}
\acrodef{ICP}{Iterative Closest Point}
\acrodef{SIFT}{Scale-Invariant Feature Transform}
\acrodef{RPE}{Relative Pose Error}
\acrodef{PoE}{Power over Ethernet}
\acrodef{TnR}{Teach-and-Repeat}
\acrodef{WILN}{Weather-Invariant Lidar-based Navigation}
\acrodef{CPU}{Central Processing Unit}
\acrodef{RAM}{Random Acces Memory}

\usepackage{fancyhdr}
\fancypagestyle{withfooter}{
  
  \fancyhead[L]{}
  \fancyhead[R]{}
  \fancyfoot[C]{\footnotesize Accepted to the IEEE ICRA Workshop on Field Robotics 2024}
}

\begin{document}
\maketitle
\thispagestyle{withfooter}
\pagestyle{withfooter}

\begin{abstract}

This report presents a wearable plug-and-play platform for data acquisition in the field.
The platform, extending a waterproof Pelican Case into a 20~kg backpack offers 5.5~hours of power autonomy, while recording data with two cameras, a lidar, an \ac{IMU}, and a \ac{GNSS} receiver.
The system only requires a single operator and is readily controlled with a built-in screen and buttons.
Due to its small footprint, it offers greater flexibility than large vehicles typically deployed in off-trail environments.
We describe the platform's design, detailing the mechanical parts, electrical components, and software stack.
We explain the system's limitations, drawing from its extensive deployment spanning over 20~kilometers of trajectories across various seasons, environments, and weather conditions.
We derive valuable lessons learned from these deployments and present several possible applications for the system.
The possible use cases consider not only academic research but also insights from consultations with our industrial partners.
The mechanical design including all CAD~files, as well as the software stack, are publicly available at \url{https://github.com/norlab-ulaval/backpack_workspace}.
\end{abstract}
\section{Introduction}
\label{sec:introduction}
\begin{figure}[tbp]
	\centering
	\includegraphics[width=0.48\textwidth]{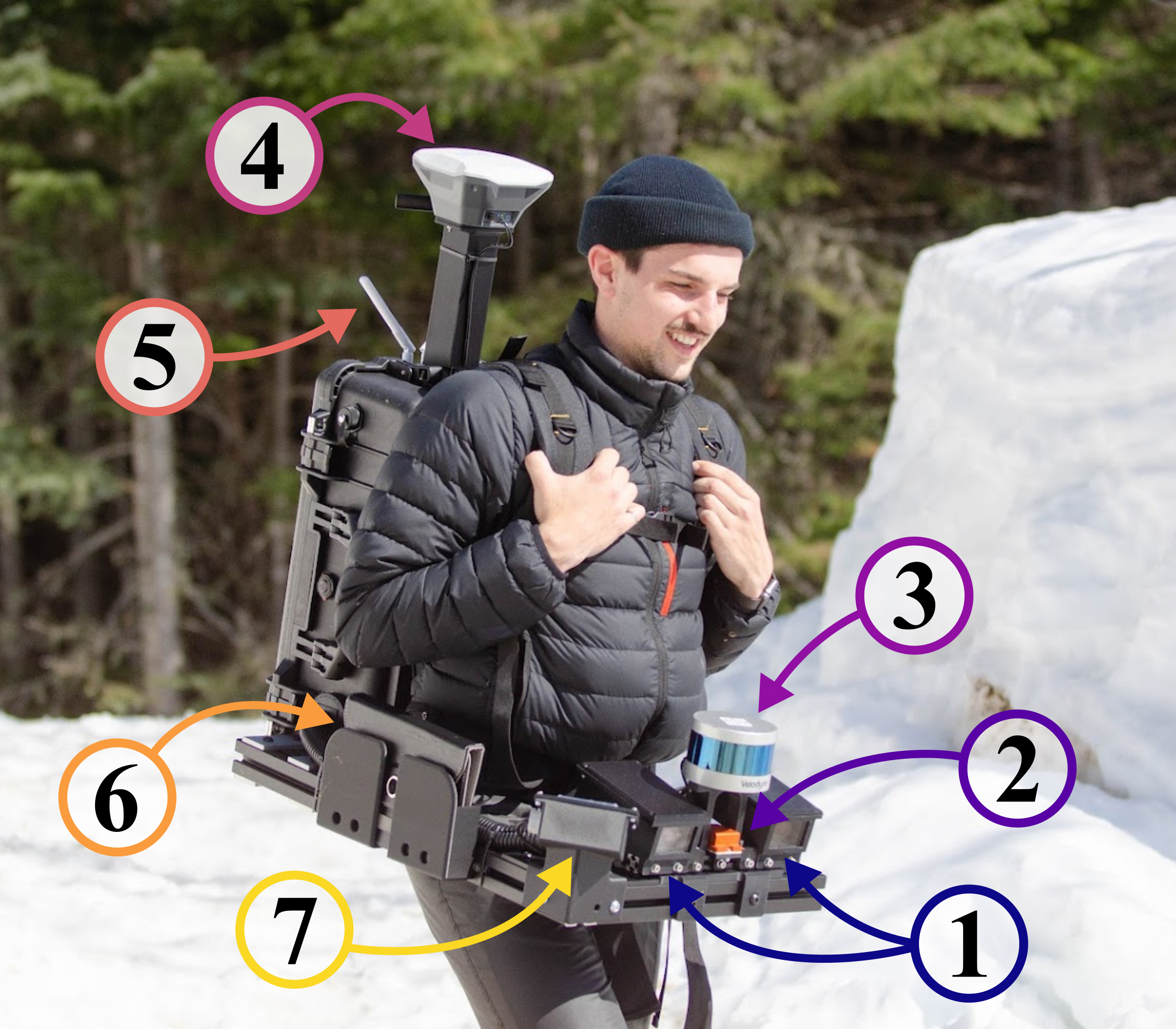}
	\caption{Picture of the developed backpack.
 Main components are identified as follows: \textbf{(1)} Two Basler a2A1920-51gcPRO cameras, \textbf{(2)} Xsens MTI-30 \ac{IMU}, \textbf{(3)} VLP16 3D lidar, \textbf{(4)} Emlid Reach RS+ GPS receiver, \textbf{(5)} Ubiquiti UniFi UAP-AC-M Wi-Fi antenna, \textbf{(6)} visualization tablet, and \textbf{(7)} control panel.}
	\label{fig:backpack}
 \vspace{-0.1in}
\end{figure}
Real-life datasets are essential to improve autonomous navigation algorithms \citep{liu2024botanicgarden}.
To capture the complexity of the world, these datasets must showcase a wide variety of environments and weather conditions. 
Several datasets acquired in urban contexts used a standard car equipped with sensors as the acquisition platform \citep{geiger2013vision, cordts2016cityscapes}.
In comparison, off-road environments require specialized ground vehicles to navigate the diversified and challenging terrains \citep{Baril2022, triest2022tartandrive}.
These platforms are usually large and powerful, which makes them hard to deploy without a large team. 
One solution to simplify off-road data acquisition is to use drones, which are easy to deploy, highly maneuverable, and independent of the terrain. 
However, their battery life is highly limited, hindering their capacity to travel long distances \citep{mozaffari2019tutorial, rohan2019advanced}.



On a different path, multiple companies started building wearable mobile solutions, such as the Leica Pegasus Backpack,\footnote{\url{https://leica-geosystems.com/products/mobile-mapping-systems/capture-platforms/leica-pegasus-backpack}} the Mosaic Xplor,\footnote{\url{https://www.mosaic51.com/cameras/mosaic-xplor/}} and the NavVIS VLX 3.\footnote{\url{https://www.navvis.com/vlx-3}}
These devices are considered top-of-the-line systems but are generally expensive and closed-source.
Any project-dependent customization, often needed for research projects, is therefore challenging.
Due to difficulties in traversing forest landscapes, the forestry research field has also proposed multiple backpack data acquisition systems.
\citet{campos2018backpack} developed a low-cost platform for navigation in forest environments.
The platform is equipped with an omnidirectional camera, in addition to an \acf{IMU} and a \acf{GNSS} antenna.
\citet{goebel2023backpack} created their version, adding a 3D lidar and a multi-spectral camera.
The two proposed designs are good examples of portable platforms but are not built for winter weather when sensors must be protected from the snow and the cold, which can reach as low as \SI{-50}{\celsius}.
Another version, designed by \citet{chahine2021mapping}, incorporated an active theodolite prism for ground truth generation from a total station.
The system was used in winter but not during snowfall, lacking the necessary protection.

In this paper, we present a technical report on the development of a cost-efficient acquisition platform, depicted in \autoref{fig:backpack}.
The developed platform is a user-friendly, water-resistant portable backpack that allows the recording of a variety of sensor data, such as images, point clouds, satellite positioning, and inertial data.
Built with a small status screen and two control buttons, the system only requires a~single operator to be deployed, without the need for an external computer for data acquisition. 
The small footprint of the acquisition platform allows for the collection of data in narrow and hard-to-access spaces for robotic vehicles.
The following sections detail the design process leading to this data-gathering platform. 
The used material is listed, followed by an analysis of the system's performances and limitations.
The final sections address the lessons learned during the process, and highlight some possible applications for research and industry sectors.

\section{Platform Details}
\label{sec:platform_details}

\begin{figure}[htbp]
    \centering
    \vspace{0.1in}
    \includegraphics[width=0.49\textwidth]{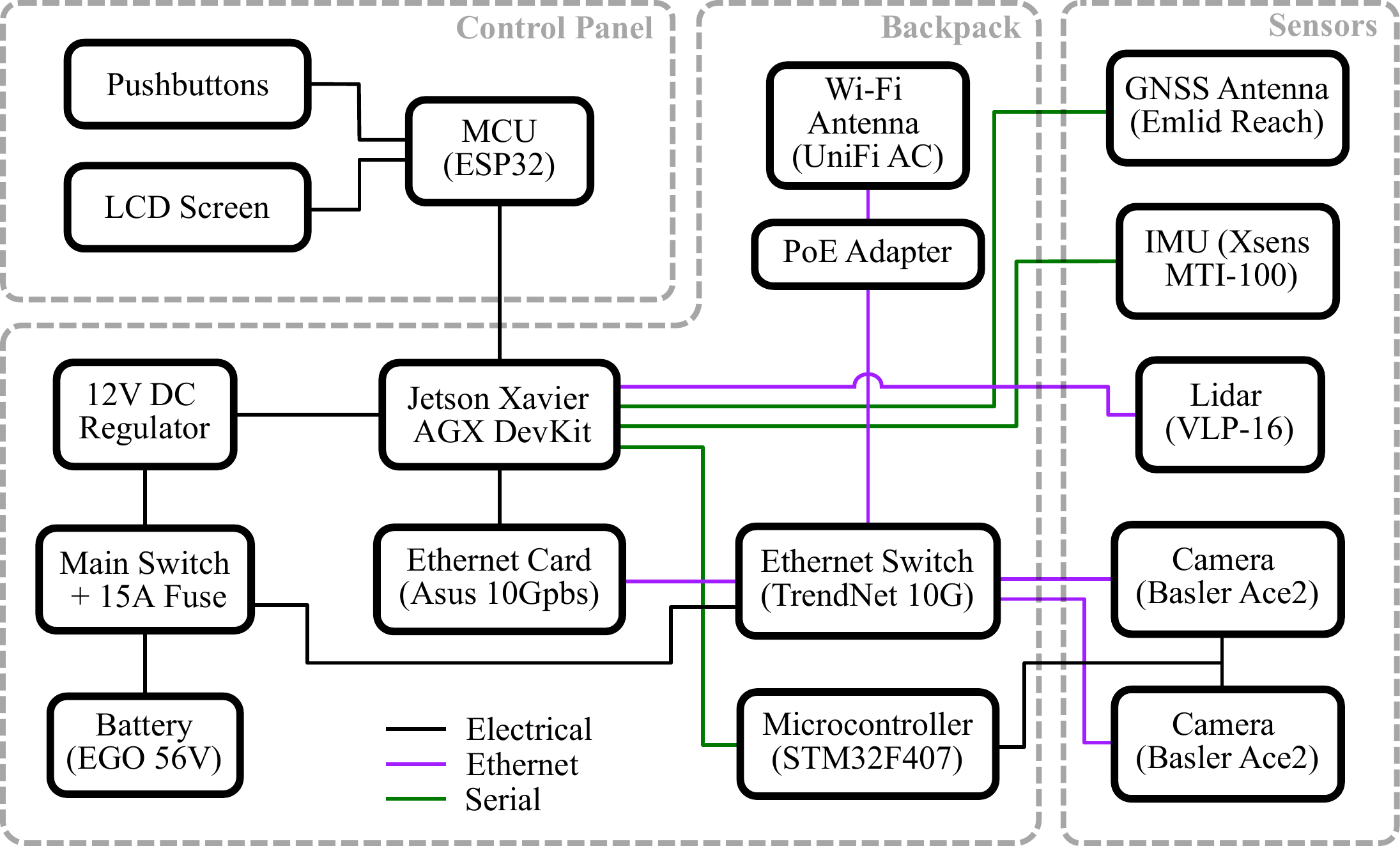}
    \caption{Block diagram of the hardware components of the mobile data acquisition system. Black lines show electrical connections, purple lines stand for Ethernet links and green lines are USB or serial cables.}
    \label{fig:schematic_hardware}
\end{figure}

The platform is composed of three main sections: the backpack itself, the sensors, and the control panel.
A block diagram of the backpack's hardware components is presented in \autoref{fig:schematic_hardware}, where the subsystems and the links between them are displayed.
The figure also shows the interface type connecting each module, \ie~electrical, Ethernet, or serial.
Note that the final choice of the presented modules was determined after three version iterations of the platform, which were intertwined by outdoor testing and multiple deployments, described in \autoref{sec:lessons_learned} and \autoref{sec:use_cases}.

\subsection{Backpack}

\begin{figure}[htbp]
    \centering
    \vspace{0.1in}
    \includegraphics[width=0.47\textwidth]{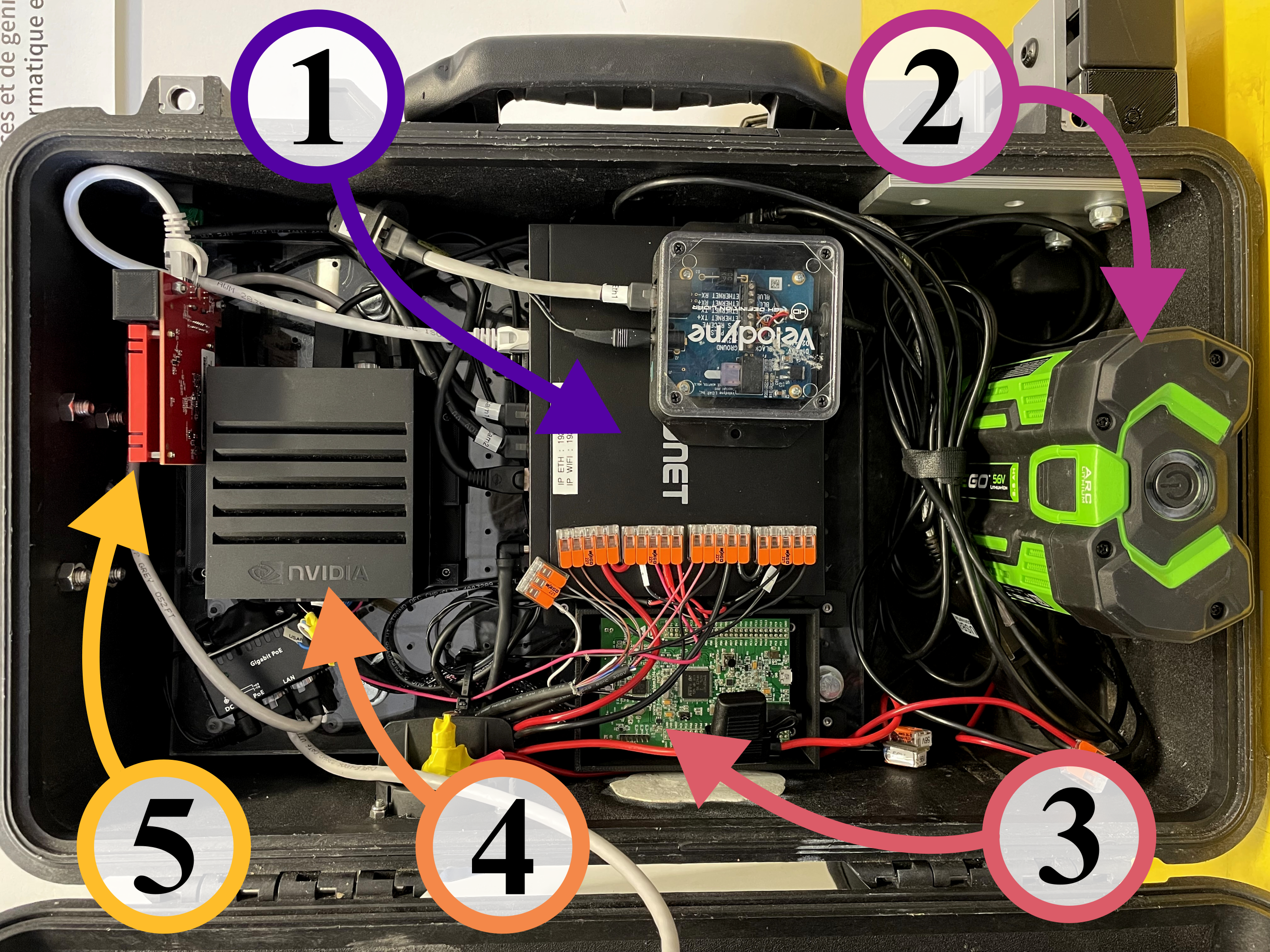}
    \caption{Picture of the inside of the backpack. Main components are identified as follows: \textbf{(1)} TRENDnet TEG-S762 switch, \textbf{(2)} EGO battery \SI{2.5}{\ampere\hour}, \textbf{(3)} STM32F407 microcontroller, \textbf{(4)} Nvidia Jetson Xavier AGX Developer Kit, and \textbf{(5)} Asus XG-C100C \SI{10}{\giga b\per\second} PCIe.}
    \label{fig:backpack_inside}
\end{figure}

The backpack's core frame is a Pelican Case 1510, which is roughly the size of a person's back and has the advantage of being water-resistant.
The main power comes from a rechargeable \SI{56}{\volt} Ego battery installed on a custom-made 3D printed connector, depicted in \autoref{fig:backpack_inside}.
Different battery formats can be used, allowing to adjust the backpack's weight between \SI{18}{\kilo\gram} and \SI{20.5}{\kilo\gram}.
The lightest battery is \SI{2.5}{\ampere\hour} and lasts around \SI{1.7}{\hour}, while the heaviest is \SI{7.5}{\ampere\hour} for a total battery life of \SI{5.5}{\hour}.
The embedded computer is a Jetson Xavier AGX Developer Kit, which records all the sensors' data directly into a 970 EVO Plus NVMe M.2 SSD of \SI{1}{\tera\byte}.
This computing unit is equipped with two embedded \SI{1}{\giga b\per\second} Ethernet interfaces.
Resulting from a lack of bandwidth that led to data losses, the computer was complemented with an Asus XG-C100C \SI{10}{\giga b\per\second} PCIe card that is connected to a TRENDnet TEG-S762 switch, offering two \SI{10}{\giga b\per\second} ports and four \SI{2.5}{\giga b\per\second} ports.
For better control over the image acquisition, an STM32F407 microcontroller acts as an external trigger to synchronize the acquisition from both cameras, but also for precise exposure time control.
The trigger signal is generated using hardware timers in the microcontroller, providing more robust timing than software-based solutions.
Finally, an antenna from Ubiquiti, the UAP-AC-M UniFi AC Mesh AP,  is connected to its own Ethernet card, and serves as a hotspot to allow wireless connection into the backpack's computer.

\subsection{Sensors}

The platform contains four main sensors.
For vision, we integrated two industrial cameras from Basler (i.e., the a2A1920-51gcPRO) in a stereo-calibrated configuration with a baseline of \SI{18}{\centi\meter} and hardware-triggered system using the microcontroller mentioned above.
Each camera captures \num{12}-bit channel images and has a \SI{1}{\giga b \per \second} bandwidth, which justifies the Ethernet card upgrade on the Jetson Xavier AGX.
In addition to the camera, the platform is equipped with a Velodyne VLP-16 lidar, which also communicates using Ethernet.
The lidar is plugged into its own Ethernet card to mitigate any risk of entanglement between signals, leading to loss of information.
Finally, the platform is equipped with an Xsens MTI-30 \ac{IMU} and an Emlid Reach RS+ \ac{GNSS} receiver.
\begin{figure*}[htbp]
	\centering
        \vspace{0.1in}
	\includegraphics[width=0.98\textwidth]{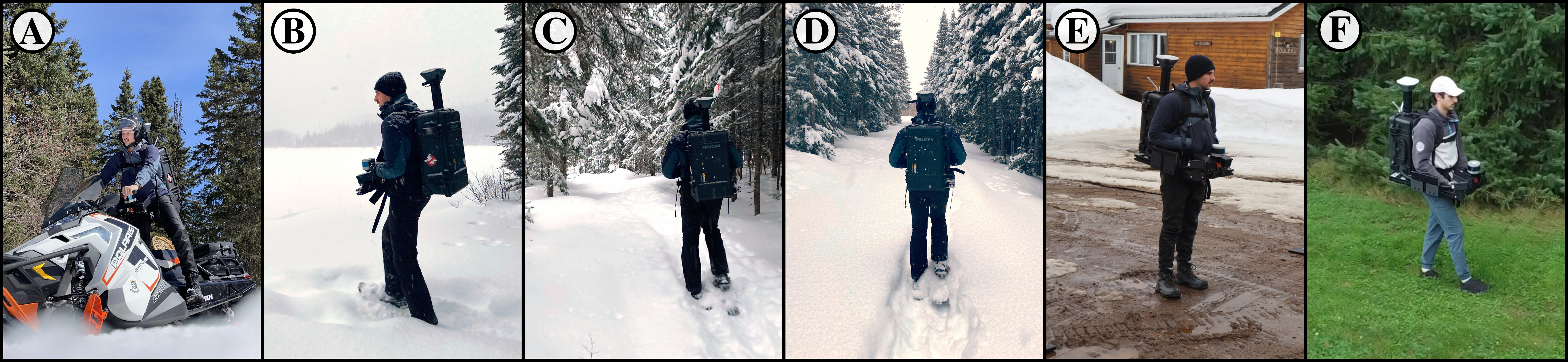}
	\caption{Examples of environments traveled with the acquisition platform. \textbf{(A)} Winter displacement on a snowmobile, \textbf{(B)} Winter frozen lake, \textbf{(C)} Winter dense forest, \textbf{(D)} Winter tree corridor, \textbf{(E)} Spring muddy forest, and \textbf{(F)} Summer forest.}
	\label{fig:mosaic}
\end{figure*}

\subsection{Control Panel}
\label{sec:platform_control_panel}

An ESP32 microcontroller communicates with the Jetson Xavier AGX over USB to display the status of each sensor to the user on an LED screen in front of the platform.
It shows four types of messages:
\begin{itemize}
    \item \texttt{OFF}: The sensor is not used.
    \item \texttt{IDLE}: The sensor is ready for data acquisition.
    \item \texttt{REC}: The sensor's data is being recorded.
    \item \texttt{ERR}: There is a problem with the sensor.
    \item \texttt{CAL}: The sensor is undergoing calibration (\ac{IMU} only).
\end{itemize}
Finally, the screen also presents the remaining free space on the drive and the cameras' acquisition rate.
The ESP32 is also connected to two push buttons, which allow to start the sensors and to enable the data recording.


\subsection{Performances Analysis}
\label{sec:platform_performances}


Based on the components installed in the platform, we analyzed the necessary bandwidth while recording data from all the sensors.
The Ethernet transmission rate is examined for each of the three Ethernet cards installed on our Jetson Xavier AGX, \ie~the two embedded interfaces and the external PCIe card.
The bandwidth coming from the camera is the highest with around \SI{872}{\mega b/\second}, while the transmission rate from the lidar is around \SI{8}{\mega b/\second}.
The last Ethernet card is plugged directly into the Ubiquiti antenna, which uses no bandwidth unless another computer is connected to it for visualization. 
An analysis of the total CPU load showed that an average of \num{5.5} cores are used from the \num{8} available, while the system uses \SI{1.55}{\giga\byte} of RAM during recording.

In relation to the performances, we estimated the energy consumption needed to power all the hardware in the platform.
The total consumption is approximately \SI{79}{\watt}, and is mainly from the computer and the Ethernet switch, using respectively \SI{40}{\watt} and \SI{12}{\watt}.

\subsection{Limitations}
\label{sec:platform_limitations}

A wearable data acquisition platform offers simplified operation, but may lack some features commonly found in standard vehicles.
For instance, no odometry data is available because of the absence of wheel encoders, which can be useful for numerous autonomous navigation algorithms.
To compensate, the \ac{IMU} is used to estimate the velocity and the position by integrating the accelerometer values.
Since these values drift quickly, the \ac{IMU}'s odometry estimations are corrected using a 3D map built from the lidar point clouds.
The code is available on GitHub.\footnote{\url{https://github.com/norlab-ulaval/imu_odom}}

Maintaining a consistent speed when gathering data is challenging due to human factors.
It is demanding to keep the same walking speed on long distances, especially if the desired speed is low.
The significant weight, comparable to a 6-year-old child, mainly comes from the enclosure and the aluminum structure. 
These respectively allow the platform to be resistant to harsh weather and vibrations, thus providing more persistent calibrations.
In relation to the weight, the platform can be challenging or impossible to wear if the operator is injured, slowing the data collection.
Lastly, oscillations are observed in the sensors' trajectories because of the walking movements, which can be troublesome for some experiments.

\label{sec:lessons_learned}
\section{Lessons Learned}

This section gives insights into the development of a robotic acquisition platform based on the problems encountered during the backpack design.
These lessons were learned after more than \SI{20}{\kilo\meter} of traveled distance with the backpack in several environment types, as depicted in \autoref{fig:mosaic}.
The system was deployed all year around, facing challenging weather, such as snowstorms, rain, and warm summer days.
The platform was also used in three industrial demonstrations, proving its robustness and reliability.

\textbf{Bandwidth} - The first purpose of the backpack was a platform to record camera images easily.
Since the two selected industrial cameras communicate at a rate of \SI{1}{\giga b \per \second}, it was challenging to obtain the maximum \ac{FPS} given in the specification sheet.
The bandwidth was first limited by the switch, which had \SI{1}{\giga b \per \second} per port. 
The flux from the two cameras being \SI{2}{\giga b \per \second}, the transfer bandwidth from the switch to the Jetson Xavier AGX was not sufficient.
The effect was worse at first since the lidar was also plugged into the same switch, then it was moved to its independent Ethernet card.
After multiple attempts to optimize the communication channels in the switch itself, we came to the conclusion that the best way to develop a robust communication link from the cameras to the main computer was to increase the hardware performance.
By adding a \SI{10}{\giga b \per \second} Ethernet link, the cameras' \ac{FPS} is maximized without any packet loss.

\textbf{Plug-and-play System} - With the intention of not having to transport a computer during data gathering, we added a small LED screen and two buttons as described in \autoref{sec:platform_control_panel}.
The screen allows for real-time updates about the sensors and the computer, giving the user certainty about the recording information.
In our case, it is mostly useful to detect malfunctions, such as disconnected sensors, cameras overheating, lidar total obstruction, and \ac{GNSS}-denied environment.
Moreover, displaying the remaining free disk space allows for better time management, especially during long deployments.
An Android tablet provides real-time visualization of the images and point clouds, through the ROS-Mobile app,\footnote{\url{https://github.com/ROS-Mobile/ROS-Mobile-Android}} which would not be possible without the high-bandwidth Wi-Fi antenna.
This setup proved beneficial to adjust camera parameters in changing environment lighting, since it allows to validate the image quality before a new recording.
From our experiences, investing time upfront in a user-friendly platform allows for faster data gathering later on, while a status display provides a more robust and efficient acquisition.

\textbf{Camera Power Supply} - For outdoor data collection, industrial cameras should be protected from rain, dust, and snow using a housing.
However, by encasing them, the cameras will often overheat due to a lack of airflow.
This effect is enlarged when using \ac{PoE} as a power supply source for the cameras, since the provided voltage is higher than the cameras' rating. 
The excess is then converted to heat, which increases the ambient temperature surrounding the camera until it hits the protection threshold, causing lag, and even shutdown.
We found that powering the camera directly from a standard power cable, instead of using the \ac{PoE}, alleviates overheating problems.

\section{Possible Applications}
\label{sec:use_cases}

\begin{figure}[htbp]
    \centering
    \vspace{0.1in}
    \includegraphics[width=0.47\textwidth]{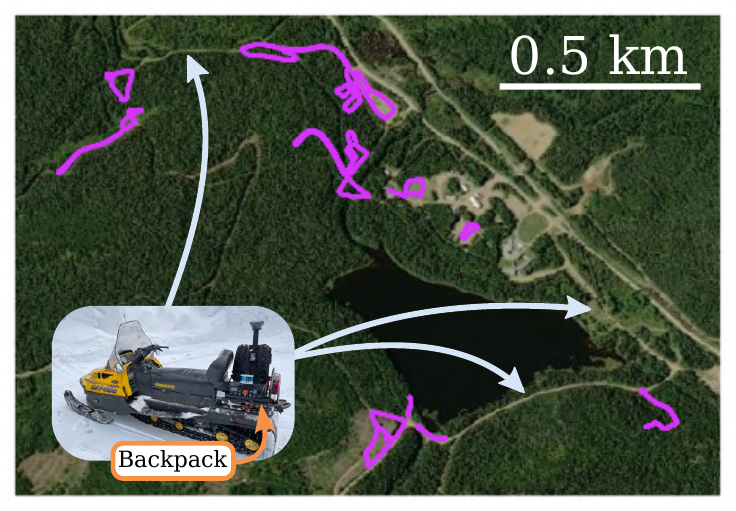}
    \caption{Satellite image of the Montmorency Forest, highlighting all the trajectories traveled on a one-day span in winter.
    The purple lines are the \ac{GNSS} positions from the \num{29} recorded trajectories, while the white arrows point to the roads traveled with the snowmobile.
    The backpack recording platform is attached to the end of the snowmobile only for the displacement between regions.}
    \label{fig:gps_trajectories}
\end{figure}

The main reason behind our platform was to develop an easy-to-deploy system, allowing to record on large territory and narrow environments, such as off-trail paths.
With our system, a single person collected the \SI{10}{\kilo\meter} BorealHDR dataset \citep{gamache2023exposing}, which was mostly recorded in winter, at the Montmorency Forest near Quebec City, Canada.
The versatility of our platform allowed us to attach it to a snowmobile, gaining speed to cover a large part of the forest, as displayed in \autoref{fig:gps_trajectories}.
In a one-day deployment, we recorded \num{29} trajectories totaling \SI{4.5}{\kilo\meter}, on an area of \SI{2.22}{\kilo\meter\squared}.
The characteristics of the platform allowed us to wear snowshoes, which were essential for navigating deep snow.
A dataset, such as BorealHDR, would have necessitated more time to collect and a larger team if it had been gathered using a standard autonomous platform.

\begin{figure}[htbp]
    \centering
    \vspace{0.1in}
    \includegraphics[width=0.47\textwidth]{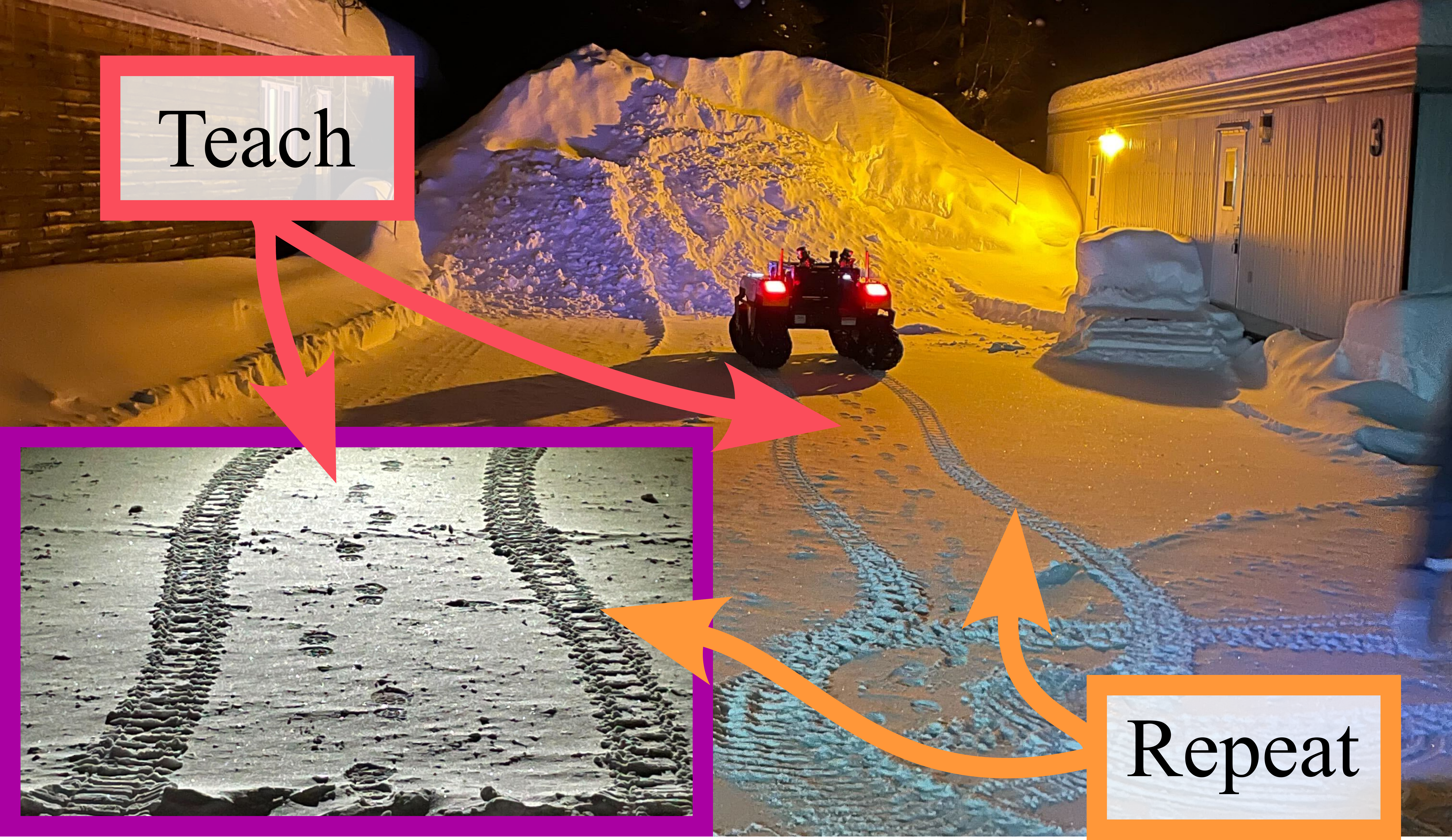}
    \caption{Example of \ac{TnR} application.
    The \textit{teach phase} was recorded using our platform and is visible by following the imprint of walking steps in the snow.
    The \textit{repeat phase} was executed with a Clearpath Warthog platform.
    The purple box highlights the \textit{repeat path} centered on the walking steps from the \textit{teach path}.}
    \label{fig:tnr}
\end{figure}

Our system was also used to perform \ac{TnR}, which consists of manually recording a trajectory and repeating it autonomously \citep{furgale2010visual}.
We used the \ac{WILN} framework,\footnote{\url{https://github.com/norlab-ulaval/wiln}} developed by \citet{Baril2022}, where our backpack system was used to record the \textit{teach} trajectory.
Then, we repeated this same trajectory with a half-ton Clearpath Warthog mobile robot.
\autoref{fig:tnr} demonstrates the successful trajectory repetition, where the footprints in the snow align with the vehicle tracks imprints.

The presented use cases could be applied in real scenarios, for instance in forest management and inventory. 
Instead of manually measuring and identifying trees, a similar solution could be used to automatically identify and register the trees with their \ac{GNSS} coordinates, without needing a specialist on-site.
Companies, such as Gaia-AI\footnote{\url{https://www.gaia-ai.eco/}} and Treeswift,\footnote{\url{https://www.treeswift.com/}} have already developed wearable platforms for such applications.
\ac{TnR} could also be useful in forestry, where foresters could first walk an uncertain path and then transfer the traveled trajectory to autonomous machinery. 
Another example would be for resupply missions, since the first crossing is always performed before needing resupplying, which means the trajectory can be recorded and sent to the main base station.

These applications are examples of the challenges that we faced with our platform, but several more applications could benefit from a versatile and easy-to-deploy system for data recording.





\section{Conclusion}
\label{sec:conclusion}

In this work, we presented our wearable acquisition platform for data recording, such as camera images, 3D lidar point clouds, inertial, and \ac{GNSS} data.
This plug-and-play system allows for easy deployments and quick data gathering due to its simple transportability.
We presented the platform by dividing it into three main components: the backpack, the sensors, and the control panel.
From the experience acquired by developing the system, we presented its limitations and the lessons learned, which we judged useful for the community.
Finally, several applications for our acquisition platform were described, giving an idea of the possibilities created by a versatile and user-friendly system.


\printbibliography

@article{goebel2023backpack,
  title={{Backpack System for Capturing 3D Point Clouds of Forests}},
  author={Goebel, Mona and Iwaszczuk, D},
  journal={ISPRS Annals of the Photogrammetry, Remote Sensing and Spatial Information Sciences},
  volume={10},
  pages={695--702},
  year={2023},
  publisher={Copernicus Publications G{\"o}ttingen, Germany}
}

@article{campos2018backpack,
  title={{A backpack-mounted omnidirectional camera with off-the-shelf navigation sensors for mobile terrestrial mapping: Development and forest application}},
  author={Campos, Mariana Batista and Tommaselli, Antonio Maria Garcia and Honkavaara, Eija and Prol, Fabricio dos Santos and Kaartinen, Harri and El Issaoui, Aimad and Hakala, Teemu},
  journal={Sensors},
  volume={18},
  number={3},
  pages={827},
  year={2018},
  publisher={MDPI}
}

@article{chahine2021mapping,
  title={Mapping in unstructured natural environment: a sensor fusion framework for wearable sensor suites},
  author={Chahine, Georges and Vaidis, Maxime and Pomerleau, Fran{\c{c}}ois and Pradalier, C{\'e}dric},
  journal={SN Applied Sciences},
  volume={3},
  pages={1--14},
  year={2021},
  publisher={Springer}
}

@article{Baril2022,
  doi = {10.55417/fr.2022050},
  url = {https://doi.org/10.55417/fr.2022050},
  year = {2022},
  month = mar,
  publisher = {Field Robotics Publication Society},
  volume = {2},
  number = {1},
  pages = {1628--1660},
  author = {Dominic Baril and Simon-Pierre Desch{\^{e}}nes and Olivier Gamache and Maxime Vaidis and Damien LaRocque and Johann Laconte and Vladim{\'{\i}}r Kubelka and Philippe Gigu{\`{e}}re and Fran{\c{c}}ois Pomerleau},
  title = {Kilometer-scale autonomous navigation in subarctic forests: challenges and lessons learned},
  journal = {Field Robotics}
}

@inproceedings{triest2022tartandrive,
  title={Tartandrive: A large-scale dataset for learning off-road dynamics models},
  author={Triest, Samuel and Sivaprakasam, Matthew and Wang, Sean J and Wang, Wenshan and Johnson, Aaron M and Scherer, Sebastian},
  booktitle={2022 International Conference on Robotics and Automation (ICRA)},
  pages={2546--2552},
  year={2022},
  organization={IEEE}
}

@article{liu2024botanicgarden,
  title={{BotanicGarden: A High-Quality Dataset for Robot Navigation in Unstructured Natural Environments}},
  author={Liu, Yuanzhi and Fu, Yujia and Qin, Minghui and Xu, Yufeng and Xu, Baoxin and Chen, Fengdong and Goossens, Bart and Sun, Poly ZH and Yu, Hongwei and Liu, Chun and others},
  journal={IEEE Robotics and Automation Letters},
  year={2024},
  publisher={IEEE}
}

@article{furgale2010visual,
  title={Visual teach and repeat for long-range rover autonomy},
  author={Furgale, Paul and Barfoot, Timothy D},
  journal={Journal of field robotics},
  volume={27},
  number={5},
  pages={534--560},
  year={2010},
  publisher={Wiley Online Library}
}

@article{geiger2013vision,
  title={{Vision meets robotics: The kitti dataset}},
  author={Geiger, Andreas and Lenz, Philip and Stiller, Christoph and Urtasun, Raquel},
  journal={The International Journal of Robotics Research},
  volume={32},
  number={11},
  pages={1231--1237},
  year={2013},
  publisher={Sage Publications Sage UK: London, England}
}

@inproceedings{cordts2016cityscapes,
  title={The cityscapes dataset for semantic urban scene understanding},
  author={Cordts, Marius and Omran, Mohamed and Ramos, Sebastian and Rehfeld, Timo and Enzweiler, Markus and Benenson, Rodrigo and Franke, Uwe and Roth, Stefan and Schiele, Bernt},
  booktitle={Proceedings of the IEEE conference on computer vision and pattern recognition},
  pages={3213--3223},
  year={2016}
}

@article{rohan2019advanced,
  title={Advanced drone battery charging system},
  author={Rohan, Ali and Rabah, Mohammed and Asghar, Furqan and Talha, Muhammad and Kim, Sung-Ho},
  journal={Journal of Electrical Engineering \& Technology},
  volume={14},
  pages={1395--1405},
  year={2019},
  publisher={Springer}
}

@article{mozaffari2019tutorial,
  title={{A tutorial on UAVs for wireless networks: Applications, challenges, and open problems}},
  author={Mozaffari, Mohammad and Saad, Walid and Bennis, Mehdi and Nam, Young-Han and Debbah, M{\'e}rouane},
  journal={IEEE communications surveys \& tutorials},
  volume={21},
  number={3},
  pages={2334--2360},
  year={2019},
  publisher={IEEE}
}

@article{gamache2023exposing,
  title={{Exposing the Unseen: Exposure Time Emulation for Offline Benchmarking of Vision Algorithms}},
  author={Gamache, Olivier and Fortin, Jean-Michel and Boxan, Mat{\v{e}}j and Pomerleau, Fran{\c{c}}ois and Gigu{\`e}re, Philippe},
  journal={arXiv preprint arXiv:2309.13139},
  year={2023}
}

\end{document}